\documentclass[letterpaper, 10 pt, conference]{ieeeconf}  

\IEEEoverridecommandlockouts                              

\usepackage{graphics} 
\usepackage{epsfig} 
\usepackage{mathptmx} 
\usepackage{times} 
\usepackage{amsmath} 
\usepackage{amssymb}  
\usepackage{cite}
\usepackage{color}
\usepackage{multirow}
\usepackage[table,xcdraw]{xcolor}
\usepackage{hyperref}

\usepackage{xcolor}
\usepackage{array}
\usepackage[table]{xcolor}
\usepackage{comment}
\usepackage{float}
\usepackage{graphicx}
\usepackage{caption}
\usepackage[skip=0cm,list=true,labelfont=it]{subcaption}
\usepackage{xspace}
\usepackage{balance}
\usepackage{multirow}
\usepackage{vcell}
\begin{document}
\title{\LARGE \bf
MARS: Multi-Scale Adaptive Robotics Vision for Underwater Object Detection and Domain Generalization
}

\author{Lyes Saad Saoud$^{*}$, Lakmal Seneviratne and Irfan Hussain$^{**}$
\thanks{}
\thanks{$^{1}$Department of Mechanical Engineering, Khalifa University, Abu Dhabi, United Arab Emirates}%
\thanks{$^{2}$Khalifa University Center for Autonomous and Robotic Systems, Khalifa University, Abu Dhabi, United Arab Emirates, P O Box 127788, Abu Dhabi, UAE.$^{*}$ Email: lyes.saoud@ku.ac.ae }%
\thanks{}
\thanks{$^{**}$ Corresponding Author, Email: irfan.hussain@ku.ac.ae}%
}

\maketitle
\thispagestyle{empty}
\pagestyle{empty}

\begin{abstract}
Underwater robotic vision encounters significant challenges, necessitating advanced solutions to enhance performance and adaptability. This paper presents MARS (Multi-Scale Adaptive Robotics Vision), a novel approach to underwater object detection tailored for diverse underwater scenarios. MARS integrates Residual Attention YOLOv3 with Domain-Adaptive Multi-Scale Attention (DAMSA) to enhance detection accuracy and adapt to different domains. During training, DAMSA introduces domain class-based attention, enabling the model to emphasize domain-specific features.
Our comprehensive evaluation across various underwater datasets demonstrates MARS's performance. On the original dataset, MARS achieves a mean Average Precision (mAP) of 58.57\%, showcasing its proficiency in detecting critical underwater objects like echinus, starfish, holothurian, scallop, and waterweeds. This capability holds promise for applications in marine robotics, marine biology research, and environmental monitoring.
Furthermore, MARS excels at mitigating domain shifts. On the augmented dataset, which incorporates all enhancements (+Domain +Residual+Channel Attention+Multi-Scale Attention), MARS achieves an mAP of 36.16\%. This result underscores its robustness and adaptability in recognizing objects and performing well across a range of underwater conditions. The source code for MARS is publicly available on GitHub at  https://github.com/LyesSaadSaoud/MARS-Object-Detection/.
\end{abstract}

\begin{keywords}
Robotic Vision, Underwater Object Detection, Domain Generalization, Multi-Scale Attention, Marine Robotics.
\end{keywords}
\begin{figure}
    \includegraphics[width=\columnwidth]{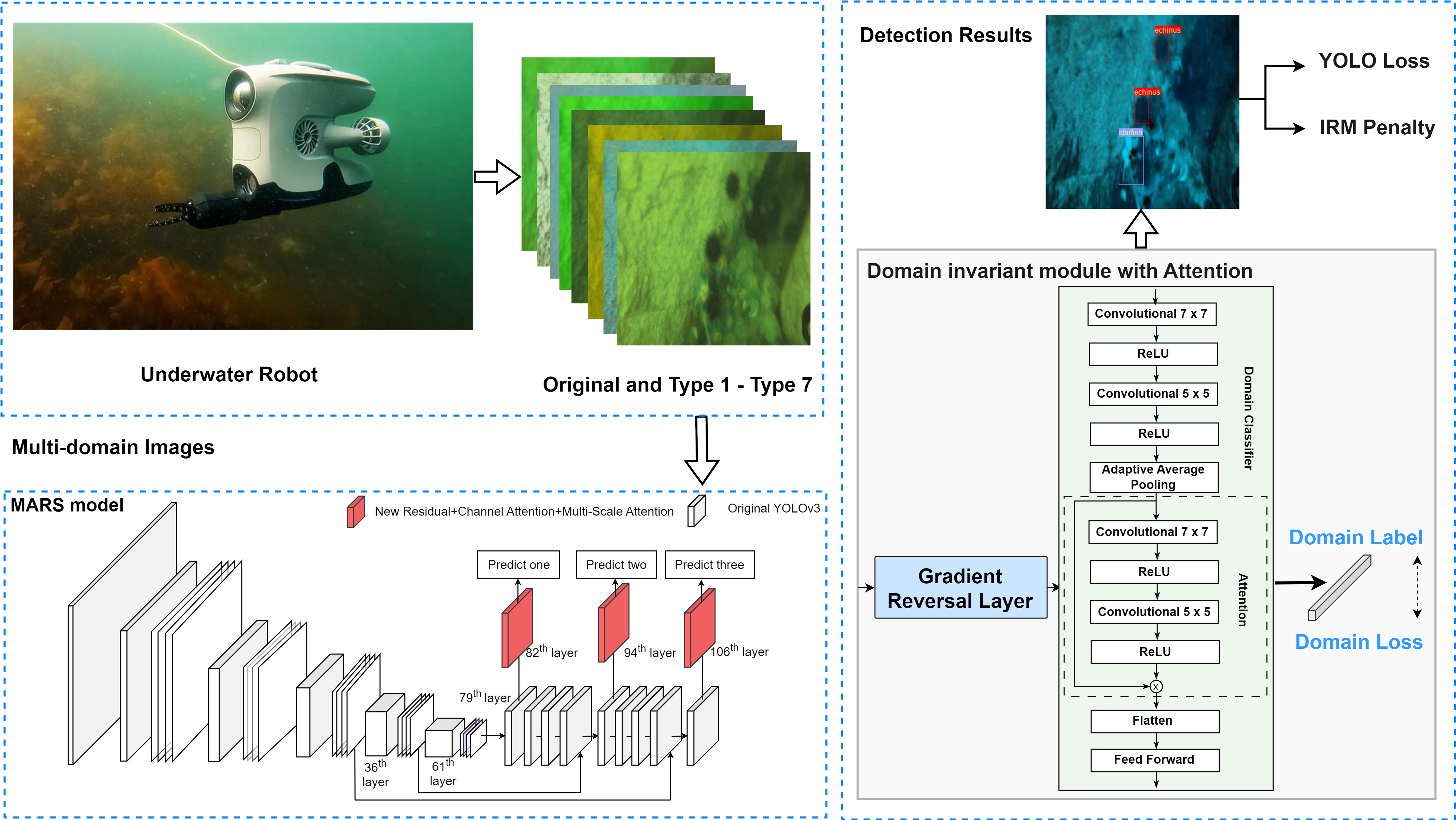}
    
    \caption{This illustration presents the architecture of the MARS (Multi-Scale Adaptive Robotics Vision) model, designed for underwater object detection. It is an extension of YOLOv3, incorporating critical components like residual layers and channel attention mechanisms integrated into specific layers ($82^{nd}$, $94^{th}$, and $106^{th}$). These additions strategically address gradient challenges during training, enhancing feature representation by focusing on informative regions while filtering out noise. The model operates as a fully convolutional network, with its foundation being the Darknet-53 backbone. It utilizes three feature maps of different scales for effective object detection. Additionally, we depict the architecture of the Domain Classifier within MARS, a vital component for domain-specific adaptation. This classifier consists of convolutional layers followed by an attention mechanism. The attention mechanism calculates weights that emphasize the most relevant features in the map, enhancing their significance. The classifier employs a softmax activation to generate domain probability distributions across predefined classes (7 classes in this example).}

    \label{fig:Image1}
\end{figure}

\section{Introduction}

Robot vision in underwater environments holds immense significance for ensuring maritime safety, environmental preservation, and the efficient management of underwater infrastructure. This critical technology aids in identifying submerged hazards such as rocks, wrecks, and coral reefs, thereby assisting vessels in navigating treacherous waters and averting collisions. Furthermore, robot vision plays an indispensable role in the maintenance of underwater infrastructure, including pipelines, cables, and offshore platforms, contributing to damage prevention and the effective management of critical assets. Additionally, it serves as a key enabler for identifying and mitigating underwater debris and pollutants, thereby safeguarding marine life.

However, despite its critical significance, robot vision in underwater environments faces formidable challenges due to inherent variations in object appearance. Factors such as viewpoint, background, lighting conditions, and image quality introduce complexity and diversity in visually representing underwater objects. These variations often lead to domain shifts, where training data distribution significantly differs from testing data. This domain shift can substantially degrade the 
performance of object detection models, limiting their effectiveness in real-world scenarios \cite{recht2019imagenet-1, hendrycks2019benchmarking-2}.

Addressing this challenge by augmenting training data, although valuable, is hindered by the labor-intensive and costly process of image annotation. Furthermore, the scarcity of labeled datasets specific to robot vision in underwater environments further compounds the challenges associated with this approach. To mitigate domain shift, domain adaptation techniques have been developed. These techniques adjust the parameters of a pre-trained model using unlabeled data from the target domain. However, securing sufficient and representative target domain data remains a persistent challenge \cite{chen2018domain-3, xu2020exploring-4, hsu2020progressive-5}.

Recent advancements in domain generalization offer a promising alternative for mitigating domain shifts in robot vision. In contrast to domain adaptation, domain generalization aims to develop a single set of model parameters that perform effectively on previously unseen domains, without the need for target domain data or a separate adaptation step \cite{li2017deeper-6}. This approach is particularly valuable as it empowers models to generalize across diverse domains, even with limited training data, making them robust in various unseen conditions.

Numerous domain generalization methods have emerged, driven by advances in computer vis
ion. These techniques encompass a wide array of approaches, including those that employ feature alignment between source domains, data augmentation-based methods for domain diversity, self-supervised training, aggregation-based methods, and meta-learning paradigms \cite{motiian2017unified, li2018domain, shankar2018generalizing, dou2019domain, d2018domain, balaji2018metareg, li2019episodic}. Despite significant progress in domain generalization, its application to robot vision in underwater environments remains relatively unexplored \cite{huang2019faster, lin2020roimix, liu2020towards, CHEN202320}.

In this study, we introduce the MARS (Multi-Scale Adaptive Robotics Vision) model, an approach tailored for robot vision in underwater object detection (Fig. \ref{fig:Image1}). At the heart of the MARS model lies the Residual Attention Block, a novel fusion of residual connections and channel attention mechanisms. This combination enhances the model's capability to capture intricate object details, emphasize pertinent features, and consequently, achieve superior detection accuracy.

\begin{figure*}[ht]
\centering
\includegraphics[width=0.9\textwidth]{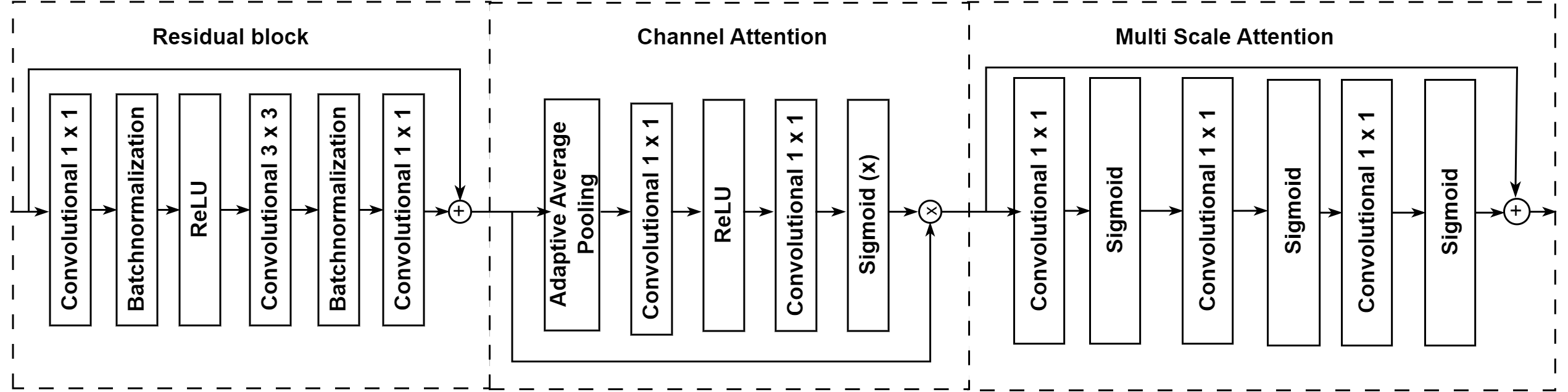}

\caption{This overview offers a glimpse into the novel components incorporated within the MARS model, an innovative approach for underwater object detection. Leveraging the YOLOv3 framework as its base, this advanced design smoothly incorporates Residual Layers, Channel Attention, and Multi-Scale Attention modules. Together, these elements enable the precise capturing of intricate object details while underscoring the importance of pertinent features.}
\label{model_architecture}
\end{figure*}

MARS combines Multi-Scale Attention, Channel Attention, and Domain Attention into a unified framework designed to address the domain generalization challenge in robot vision for underwater object detection. Our contributions are outlined as follows:

\begin{itemize}

\item \textbf{Multi-Scale Attention Fusion}: At the core of MARS is the Multi-Scale Attention Fusion module. This module seamlessly integrates Multi-Scale Attention, Channel Attention, and Domain Attention, allowing the model to capture object-specific details, amplify critical features, and improve detection precision.

\item \textbf{Seamless Adaptation to Varied Underwater Domains}: MARS stands out for its natural adaptability to the unique characteristics of diverse underwater domains, all without the need for extensive domain-specific training data. MARS excels in detecting objects across a range of underwater conditions, showcasing its exceptional versatility.

\item \textbf{Augmented Object Discrimination}: An essential feature of MARS is its incorporation of adaptive object discrimination mechanisms. Through precise recognition of domain-specific cues during object detection, MARS significantly enhances object discrimination and detection performance. This adaptiveness is a distinguishing aspect of MARS, setting it apart from previous approaches.
\end{itemize}

Collectively, these contributions position MARS as a capable solution for robust underwater object detection. It effectively tackles the intricate challenges posed by domain shifts in underwater environments while maintaining a focus on technical rigor and originality.

\section{Related Works}

Underwater object detection has seen various approaches to address domain shifts. Techniques such as RoIMix, which introduces Mixup at the Region of Interest (RoI) level to simulate occlusion conditions, and the use of dilated convolutions to enhance feature extraction have been explored \cite{lin2020roimix, fan2020dual, chen2020c}. While attention-based and feature-pyramid-based methodologies have contributed to feature extraction, they may still have limitations in fully mitigating domain shift challenges \cite{liang2022excavating, zhao2021composited}.

DG-YOLO, a notable solution tailored for addressing underwater domain shift through domain generalization, employs techniques such as Water Quality Transfer to expand the dataset's scale and domain diversity \cite{liu2020towards}. This approach achieves domain-invariant detection performance through domain adversarial training and invariant risk minimization.

In our prior work, we introduced a modified YOLOv3 architecture explicitly tailored for precise underwater object detection. This architecture efficiently captures intricate details and adeptly handles underwater-specific variations, resulting in significant improvements in detection accuracy.

Domain Adaptation and Generalization are critical tasks for achieving robust performance in a target domain using source domain data. In domain adaptation, the source and target domains share an identical label space. Techniques such as Domain-Adversarial Neural Networks (DANN) aim to align features via adversarial training \cite{ganin2015unsupervised}. Other strategies involve generating target domain data from source domain data using Variational Autoencoder (VAE) and Generative Adversarial Network (GAN) approaches, with extensions tailored to object detection \cite{xu2019adversarial, gong2019dlow, chen2018domain}.

Conversely, domain generalization involves training on multiple source domains and evaluating on an unseen but related target domain. Strategies include feature alignment between source domains, data augmentation techniques, self-supervised training, aggregation-based methods, and meta-learning approaches \cite{motiian2017unified, li2018domain, shankar2018generalizing, dou2019domain, d2018domain, balaji2018metareg, li2019episodic}. Despite progress in domain generalization, its application to underwater object detection remains relatively unexplored \cite{huang2019faster, lin2020roimix, liu2020towards}.

Additionally, recent developments in underwater robotics have expanded the capabilities of autonomous underwater vehicles (AUVs). Liu et al. introduced a novel method for self-supervised underwater monocular depth estimation using a single-beam echosounder (SBES) and presented a synthetic dataset for underwater depth estimation \cite{10161439}. Chen et al. designed an underwater jet-propulsion soft robot with high flexibility driven by water hydraulics, inspired by octopus propulsion \cite{10160331}. Yao et al. presented an Image-Based Visual Servoing (IBVS) leader-follower control system for heterogeneous aquatic robots \cite{10160853}. Girdhar et al. introduced CUREE, a curious underwater robot for ecosystem exploration, providing unique capabilities for underwater ecosystem studies \cite{10161282}.

In this study, we introduce MARS, a Multi-Scale Adaptive Robotics Vision model. MARS leverages Multi-Scale Attention, Channel Attention, and Domain Attention to address domain generalization challenges. At the core of MARS is the Multi-Scale Attention Fusion module, designed to capture object-specific details and enhance detection precision.

\section{The Proposed MARS Model}

The MARS (Multi-Scale Adaptive Robotics Vision) model represents a significant advancement in the field of underwater object detection, building upon the widely adopted YOLOv3 architecture \cite{redmon2018yolov3}. Tailored specifically for underwater scenarios, MARS incorporates several pivotal modifications to elevate its performance.
The MARS (Multi-Scale Adaptive Robotics Vision) model signifies a substantial leap in the realm of underwater object detection. It builds upon the widely embraced YOLOv3 architecture \cite{redmon2018yolov3} but tailors its core to excel in underwater scenarios. MARS incorporates a series of transformative adjustments to elevate its performance.

\subsection{Model Architecture}

At its heart, MARS constitutes a fully convolutional network that boasts a Darknet-53 backbone, setting it apart from conventional models, which tend to incorporate pooling layers. The unique Darknet-53 design encompasses four distinct types of residual units, each comprising a sequence of 1x1 and 3x3 convolutional layers. This innovative approach empowers the network to dynamically adapt the convolution kernel's stride during forward propagation, ultimately resulting in feature maps that are a mere 1/25 of the original input size. Consequently, the tensor size undergoes a significant reduction, rendering MARS exceptionally efficient.

\subsubsection{Multi-Scale Attention}

To augment the feature extraction process and facilitate robust underwater object detection, MARS thoughtfully integrates Multi-Scale Attention modules.

The Multi-Scale Attention module plays a pivotal role within the MARS model by meticulously focusing on informative regions within the feature maps while effectively suppressing irrelevant content. This attention mechanism seamlessly operates across different scales, profoundly enhancing the model's ability to attend to pertinent information, thereby elevating object localization and detection accuracy. The Multi-Scale Attention module is composed of three convolutional layers, each with a 1x1 kernel, subsequently followed by three Sigmoid activation functions. The Sigmoid activation scales the attention weights within the range of 0 to 1, thus enabling the module to judiciously control the significance of each spatial location within the feature map.
\subsection{Enhancements for Underwater Object Detection}

To address gradient-related challenges in deep networks and further boost the model's performance, MARS incorporates Residual Layers and Channel Attention modules.
\subsubsection{Residual Layers}

Strategically placed after each feature map layer (e.g., $82^{nd}$, $94^{th}$, and $106^{th}$ layers), Residual layers emerge as a potent tool to mitigate gradient vanishing and explosion issues. Each Residual Block consists of two Conv2d 1x1 layers, two batch normalizations, and a ReLU activation function. These Residual layers serve as invaluable shortcuts, facilitating smooth gradient flow, thereby simplifying the optimization process. Their inclusion empowers the model to effectively train deeper networks, ultimately leading to superior model performance and precise detection of underwater objects.

\subsubsection{Channel Attention}

The Channel Attention module assumes a pivotal role in extracting features across diverse scales within the MARS model. Its incorporation significantly bolsters MARS's ability to capture intricate details and emphasize relevant features, culminating in heightened accuracy and robustness in object detection. This module incorporates adaptive average pooling, two Conv2d 1x1 layers, a ReLU function, and a Sigmoid activation function, intelligently directing attention toward crucial channel-wise information.

\subsection{Domain Classifier for Robustness}

Within the intricate domain of underwater feature detection and identification, the task assumes formidable complexity due to the distortions introduced by the fluid medium, impacting both the objects and their surroundings. To fortify the algorithm's resilience in the face of underwater conditions, the MARS model seamlessly incorporates a Domain Classifier module. This module aspires to enhance the domain classification performance of YOLOv3 at each feature map layer.

The Domain Classifier comprises two Conv2d layers (7x7 and 5x5, respectively), two ReLU functions, and an adaptive average pooling layer. The output of this process then traverses an attention block, encompassing two Conv2d layers (7x7 and 5x5, respectively) and two ReLU functions. The results extracted from the adaptive average pooling and attention block undergo multiplication and flattening before being channeled into a feed-forward block.

Through the introduction of these innovative modules, including Multi-Scale Attention, Channel Attention, Residual layers, and the Domain Classifier, thoughtfully interwoven into feature map layers, MARS stands as a beacon of superior object detection in challenging underwater conditions. It serves as a specialized tool, meticulously tailored for demanding underwater applications, deftly addressing the intricate challenges inherent to underwater environments.

\section{Results}
\begin{table*}[ht]
\caption{Results for Models on Original Data without using Domain, with the values in bold showing the best-obtained values.}
\label{tab:results_original_no_domain}
\centering
\begin{tabular}{|l|c|c|c|c|c|c|}
\hline
\textbf{Model} & \textbf{Echinus} & \textbf{Starfish} & \textbf{Holoth.} & \textbf{Scallop} & \textbf{Waterweed} & \textbf{mAP} \\
\hline
Baseline (YOLOv3) & 83.67 & 71.87 & 51.32 & 64.54 & 0.00 & 54.28 \\
+Residual & 80.28 & 64.24 & 47.31 & 50.78 & 0.00 & 49.12 \\
+Channel Attention  & 82.64 & 74.60 & 55.39 & 68.40 & 0.00 & 56.21 \\
+Residual Attention & 82.96 & 74.51 & 54.10 & \textbf{69.12} & \textbf{4.76} & \textbf{57.09} \\
+Multi-Scale Attention & 56.24 & 54.36 & 20.76 & 28.17 & 0.00 & 31.91 \\
+Residual+Multi-Scale Attention & 48.12 & 58.74 & 37.27 & 29.35 & 0.00 & 34.70 \\
+Channel Attention+Multi-Scale Attention & \textbf{84.95} & 74.78 & \textbf{56.21} & 61.89 & 0.00 & 55.57 \\
+Residual+Channel Attention+Multi-Scale Attention & 80.23 & \textbf{75.56} & 53.85 & 60.74 & 2.72 & 54.62 \\
\hline
\end{tabular}
\end{table*}
\begin{table*}[ht]
\caption{Detection results for different models using augmented datasets without using Domain, with the values in bold showing the best-obtained values.}
\label{tab:results_augmented_no_domain}
\centering
\begin{tabular}{|l|c|c|c|c|c|c|}
\hline
\textbf{Model} & \textbf{Echinus} & \textbf{Starfish} & \textbf{Holoth.} & \textbf{Scallop} & \textbf{Waterweed} & \textbf{mAP} \\
\hline
Baseline (YOLOv3) & 69.08 & 20.18 & 31.20 & \textbf{44.58} & 0.00 & 33.01 \\
+Residual & 63.95 & \textbf{32.50} & 32.52 & 40.10 & 0.00 & 33.81 \\
+Channel Attention & 60.20 & 24.69 & 24.54 & 35.60 & 0.00 & 29.01 \\
+Residual Attention & 66.31 & 27.39 & 32.46 & 34.13 & 0.00 & 32.06 \\
+Multi-Scale Attention & 48.21 & 7.54 & 16.18 & 16.38 & 0.00 & 17.66 \\
+Residual+Multi-Scale Attention & 38.73 & 18.15 & 20.98 & 18.92 & 0.00 & 19.36 \\
+Channel Attention+Multi-Scale Attention & \textbf{70.19} & 21.92 & 26.41 & 31.73 & 0.00 & 30.05 \\
+Residual+Channel Attention+Multi-Scale Attention & 66.16 & 31.10 & \textbf{34.05} & 42.06 & \textbf{0.53} & \textbf{34.78} \\
\hline
\end{tabular}
\end{table*}
\begin{table*}[ht]
\caption{Detection results for different models using original datasets with Domain, with the values in bold showing the best-obtained values.}
\label{tab:results_original_with_domain}
\centering
\begin{tabular}{|l|c|c|c|c|c|c|}
\hline
\textbf{Model} & \textbf{Echinus} & \textbf{Starfish} & \textbf{Holoth.} & \textbf{Scallop} & \textbf{Waterweed} & \textbf{mAP} \\
\hline
Baseline (YOLOv3)                & 83.67          & 71.87          & 51.32          & 64.54          & 0.00          & 54.28          \\

+Domain                     & 82.04          & \textbf{77.66} & \textbf{59.86} & 62.10          & 0.00          & 56.33          \\
+Domain +Residual           & 82.49          & 69.62          & 52.17          & 64.30          & 0.00          & 53.71          \\
+Domain +Channel Attention  & \textbf{84.88} & 74.95          & 52.95          & \textbf{67.61}          & 0.68          & 56.21          \\
+Domain +Residual Attention & 82.72          & 69.96          & 52.15          & 65.61          & 0.00          & 54.09          \\ 

+Domain +Multi-Scale Attention & 73.28 & 73.08 & 47.81 & 54.73 & 0.00 & 49.78 \\
+Domain +Residual+Multi-Scale Attention & 77.74 & 70.24 & 49.39 & 56.47 & 0.00 & 50.77 \\

+Domain +Channel Attention+Multi-Scale Attention & 81.45 & 74.29 & 56.72 & 59.75 & 0.00 & 54.44 \\
+Domain +Residual+Channel Attention+Multi-Scale Attention & 84.77 & 75.34 & 57.12 & 72.28 & \textbf{3.35} & \textbf{58.57} \\
\hline
\end{tabular}
\end{table*}
\begin{table*}[ht]
\caption{Detection results for different models using augmented datasets with Domain, with the values in bold showing the best-obtained values.}
\label{tab:results_augmented_with_domain}
\centering
\begin{tabular}{|l|c|c|c|c|c|c|}
\hline
\textbf{Model} & \textbf{Echinus} & \textbf{Starfish} & \textbf{Holoth.} & \textbf{Scallop} & \textbf{Waterweed} & \textbf{mAP} \\
\hline
Baseline (YOLOv3)                & 69.08          & 20.18          & 31.20          & 44.58 & 0.00          & 33.01          \\

+Domain                      & 58.97          & 31.61          & 28.30          & 37.24          & 0.00          & 31.22          \\
YOLO3+Domain +Residual           & \textbf{71.61} & 18.54          & 30.05          & 45.07          & 0.00          & 33.05          \\
+Domain+Channel Attention n  & 65.21          & 28.35          & 30.12          & 41.79          & 0.00          & 33.09          \\
+Domain +Residual Attention & 67.97          & \textbf{38.91} & 34.28 & 38.33          & 0.00          & 35.90 \\ 
+Domain +Multi-Scale Attention & 65.94 & 24.09 & 31.17 & 38.53 & 0.00 & 31.95 \\
+Domain +Residual+Multi-Scale Attention & 65.81 & 16.97 & 25.72 & 31.39 & 0.00 & 27.98 \\
+Domain +Channel Attention+Multi-Scale Attention & 70.40 & 26.97 & 33.85 & 38.13 & 0.00 & 33.87 \\
+Domain +Residual+Channel Attention+Multi-Scale Attention & 65.38 & 27.69 & \textbf{35.90} & \textbf{50.60} & \textbf{1.22} & \textbf{36.16} \\
\hline
\end{tabular}
\end{table*}
\begin{figure*}
\begin{subfigure}{.12\textwidth}
   \includegraphics[width=\textwidth]{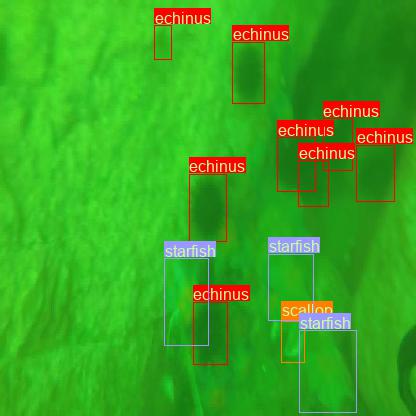}
   \caption{}
\end{subfigure}
\begin{subfigure}{.12\textwidth}
   \includegraphics[width=\textwidth]{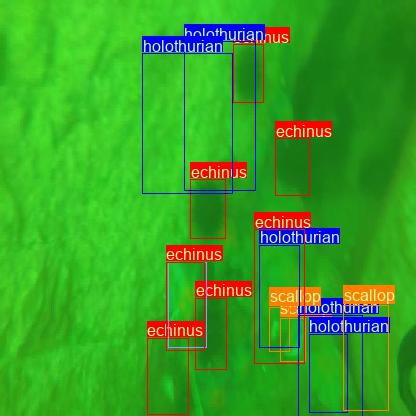}
   \caption{}
\end{subfigure}
\begin{subfigure}{.12\textwidth}
   \includegraphics[width=\textwidth]{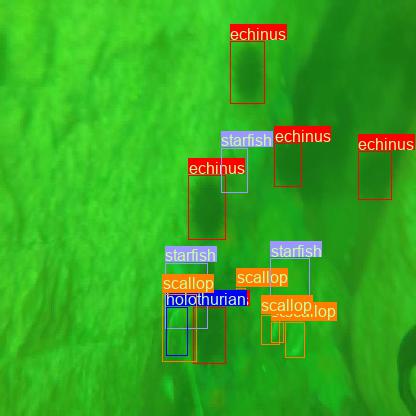}
   \caption{}
\end{subfigure}
\begin{subfigure}{.12\textwidth}
   \includegraphics[width=\textwidth]{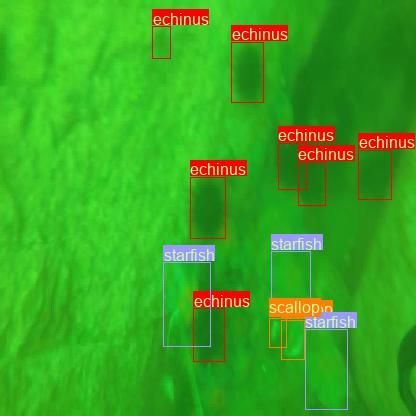}
   \caption{}
\end{subfigure}
\begin{subfigure}{.12\textwidth}
   \includegraphics[width=\textwidth]{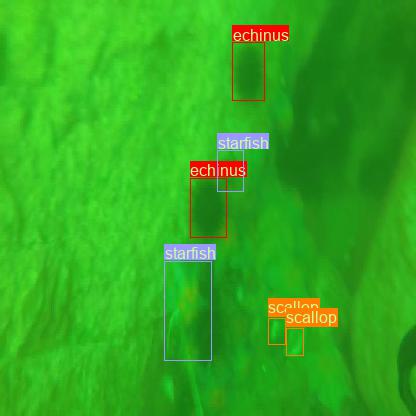}
   \caption{}
\end{subfigure}
\begin{subfigure}{.12\textwidth}
   \includegraphics[width=\textwidth]{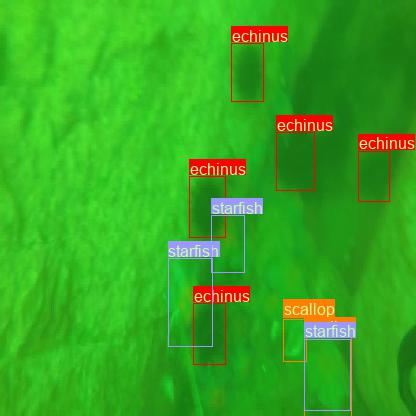}
   \caption{}
\end{subfigure}
\begin{subfigure}{.12\textwidth}
   \includegraphics[width=\textwidth]{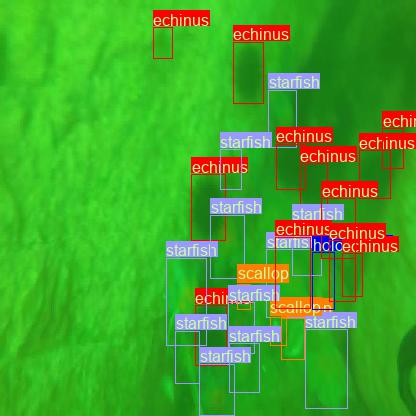}
   \caption{}
\end{subfigure}
\begin{subfigure}{.12\textwidth}
   \includegraphics[width=\textwidth]{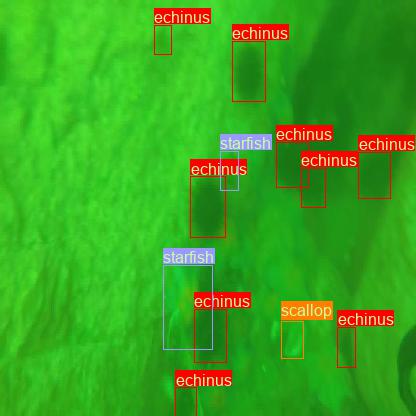}
   \caption{}
\end{subfigure}
\vfill
\begin{subfigure}{.12\textwidth}
   \includegraphics[width=\textwidth]{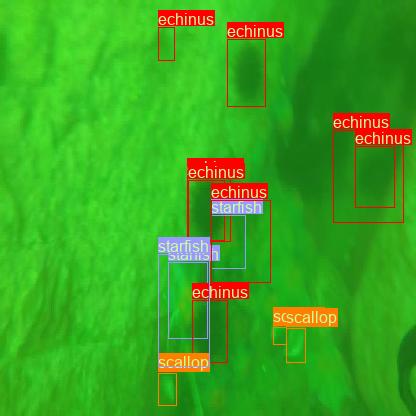}
   \caption{Model 1}
\end{subfigure}
\begin{subfigure}{.12\textwidth}
   \includegraphics[width=\textwidth]{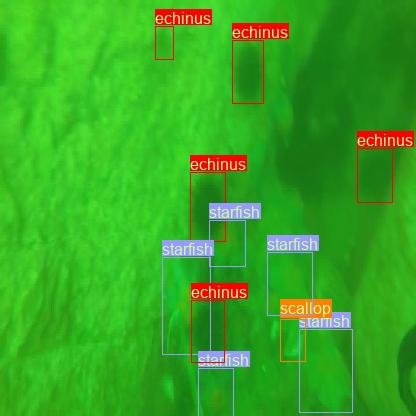}
   \caption{Model 2}
\end{subfigure}
\begin{subfigure}{.12\textwidth}
   \includegraphics[width=\textwidth]{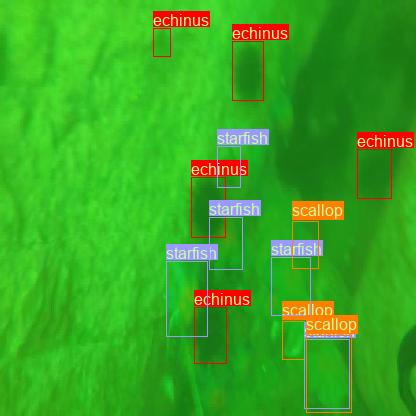}
   \caption{Model 3}
\end{subfigure}
\begin{subfigure}{.12\textwidth}
   \includegraphics[width=\textwidth]{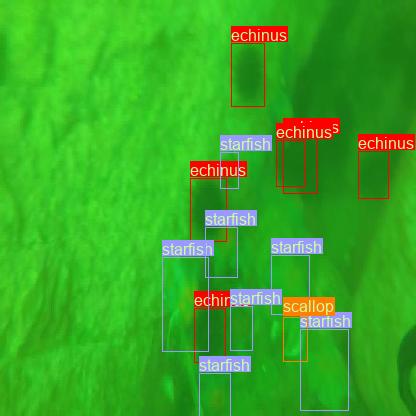}
   \caption{Model 4}
\end{subfigure}
\begin{subfigure}{.12\textwidth}
   \includegraphics[width=\textwidth]{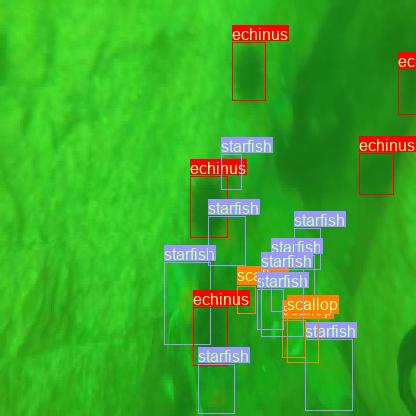}
   \caption{Model 5}
\end{subfigure}
\begin{subfigure}{.12\textwidth}
   \includegraphics[width=\textwidth]{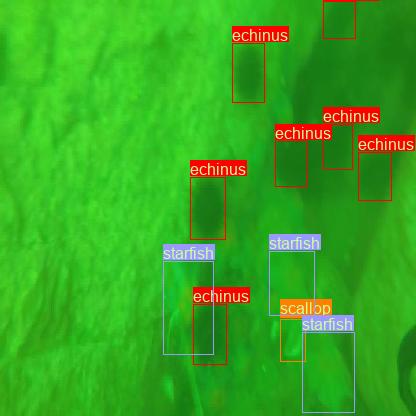}
   \caption{Model 6}
\end{subfigure}
\begin{subfigure}{.12\textwidth}
   \includegraphics[width=\textwidth]{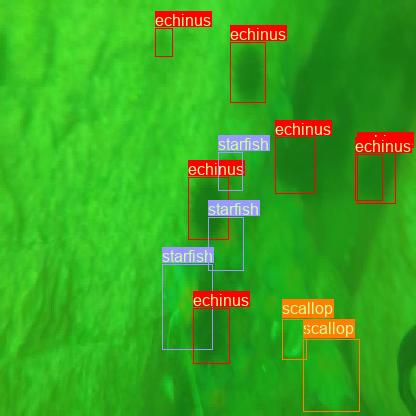}
   \caption{Model 7}
\end{subfigure}
\begin{subfigure}{.12\textwidth}
   \includegraphics[width=\textwidth]{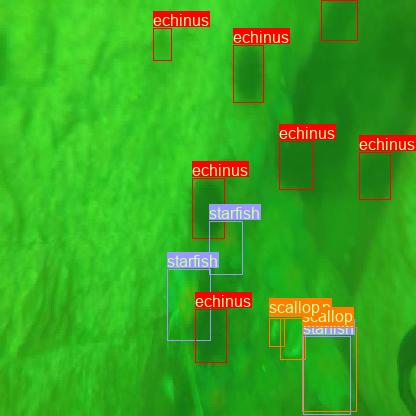}
   \caption{Model 8}
\end{subfigure}

 \caption{Performance comparison for the obtained results using the original dataset: The models from 1 to 8 are as follows: Baseline (YOLOv3),
+Residual,
+Channel Attention,
+Residual Attention,
+Multi-Scale Attention,
+Residual+Multi-Scale Attention,
+Channel Attention+Multi-Scale Attention, 
+Residual+Channel Attention+Multi-Scale Attention, respectively.}

\label{fig:1}
\end{figure*}
\begin{figure*}
\begin{subfigure}{.12\textwidth}
   \includegraphics[width=\textwidth]{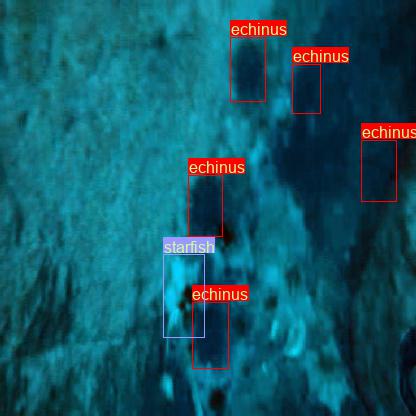}
   \caption{}
\end{subfigure}
\begin{subfigure}{.12\textwidth}
   \includegraphics[width=\textwidth]{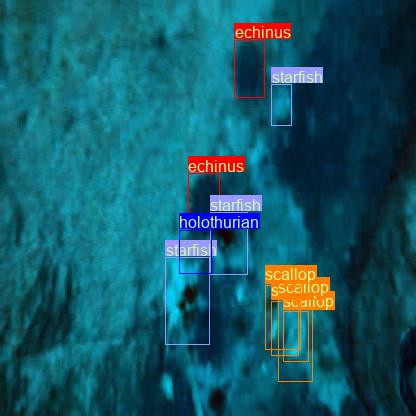}
   \caption{}
\end{subfigure}
\begin{subfigure}{.12\textwidth}
   \includegraphics[width=\textwidth]{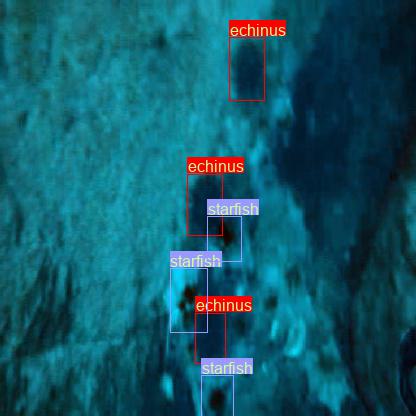}
   \caption{}
\end{subfigure}
\begin{subfigure}{.12\textwidth}
   \includegraphics[width=\textwidth]{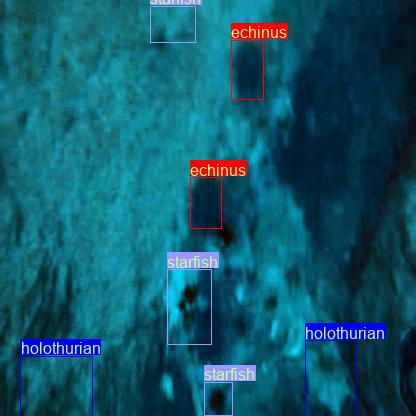}
   \caption{}
\end{subfigure}
\begin{subfigure}{.12\textwidth}
   \includegraphics[width=\textwidth]{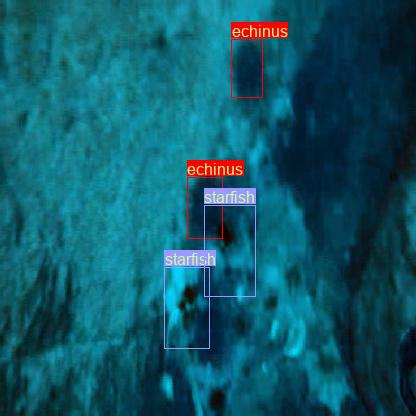}
   \caption{}
\end{subfigure}
\begin{subfigure}{.12\textwidth}
   \includegraphics[width=\textwidth]{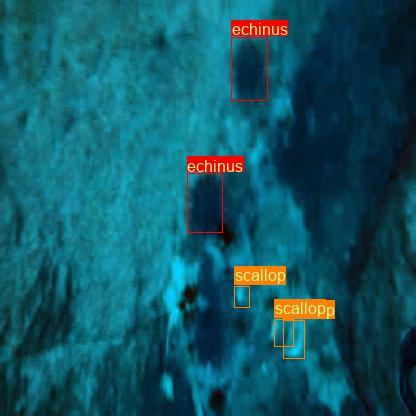}
   \caption{}
\end{subfigure}
\begin{subfigure}{.12\textwidth}
   \includegraphics[width=\textwidth]{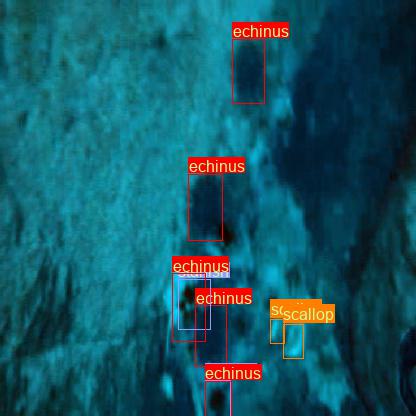}
   \caption{}
\end{subfigure}
\begin{subfigure}{.12\textwidth}
   \includegraphics[width=\textwidth]{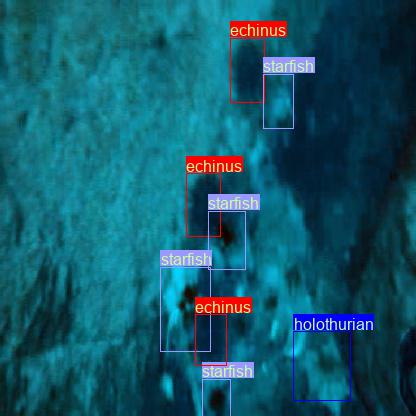}
   \caption{}
\end{subfigure}
\vfill
\begin{subfigure}{.12\textwidth}
   \includegraphics[width=\textwidth]{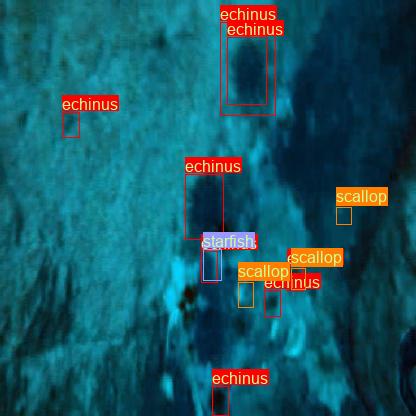}
   \caption{Model 1}
\end{subfigure}
\begin{subfigure}{.12\textwidth}
   \includegraphics[width=\textwidth]{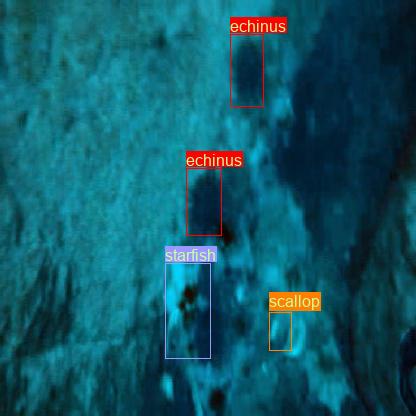}
   \caption{Model 2}
\end{subfigure}
\begin{subfigure}{.12\textwidth}
   \includegraphics[width=\textwidth]{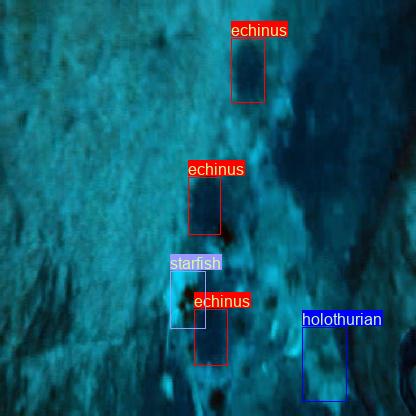}
   \caption{Model 3}
\end{subfigure}
\begin{subfigure}{.12\textwidth}
   \includegraphics[width=\textwidth]{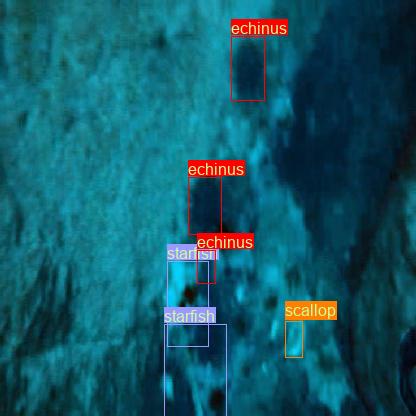}
   \caption{Model 4}
\end{subfigure}
\begin{subfigure}{.12\textwidth}
   \includegraphics[width=\textwidth]{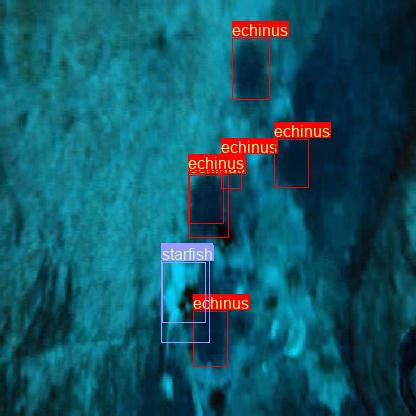}
   \caption{Model 5}
\end{subfigure}
\begin{subfigure}{.12\textwidth}
   \includegraphics[width=\textwidth]{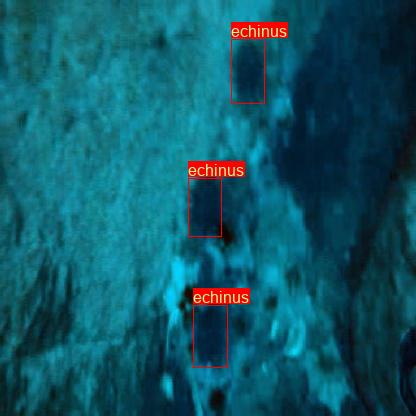}
   \caption{Model 6}
\end{subfigure}
\begin{subfigure}{.12\textwidth}
   \includegraphics[width=\textwidth]{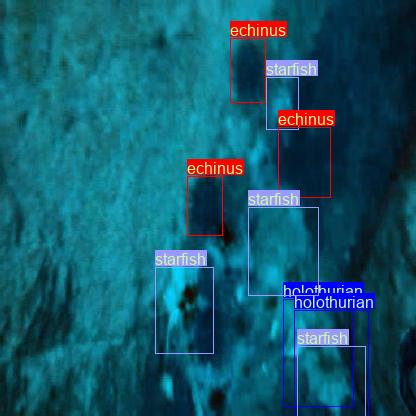}
   \caption{Model 7}
\end{subfigure}
\begin{subfigure}{.12\textwidth}
   \includegraphics[width=\textwidth]{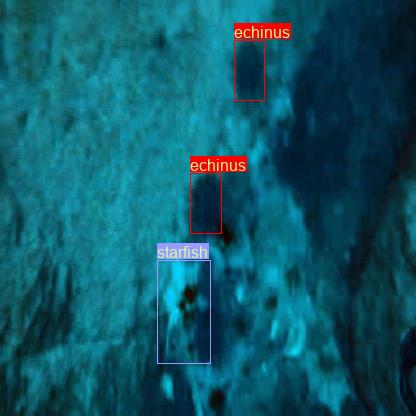}
   \caption{Model 8}
\end{subfigure}
\caption{Performance comparison for the obtained results using the augmented dataset (Type 8): The models from 1 to 8 are as follows: Baseline (YOLOv3),
+Residual,
+Channel Attention,
+Residual Attention,
+Multi-Scale Attention,
+Residual+Multi-Scale Attention,
+Channel Attention+Multi-Scale Attention, 
+Residual+Channel Attention+Multi-Scale Attention, respectively.}

\label{fig:2}
\end{figure*}
We conducted a comprehensive evaluation of the proposed MARS model using the publicly available Underwater Robot Picking Contest 2019 (URPC2019) dataset\footnote{Available at: http://en.urpc.org.cn}. This dataset comprises 3,765 training samples and 942 validation samples, categorized into five classes: echinus, starfish, holothurian, scallop, and waterweeds. Our primary objective was to assess MARS's performance in underwater object detection compared to the original YOLOv3 model. The results for each model are presented in Tables \ref{tab:results_original_no_domain}, \ref{tab:results_augmented_no_domain}, \ref{tab:results_original_with_domain}, and \ref{tab:results_augmented_with_domain}.

During the experiments, we utilized a high-performance workstation equipped with an NVIDIA GeForce RTX 4090 GPU and the PyTorch 1.13.1 framework for model training. To maintain consistency, all images were resized to a width of 416 pixels. The training process employed the Adam optimizer with a fixed learning rate of $10^{-3}$, a batch size of 32, and ran for 300 epochs. These settings were consistently applied across all models in this study.

Table \ref{tab:results_original_no_domain} presents results for models trained on the original dataset without domain-specific information. While the Baseline (YOLOv3) performs well, improvements are needed for specific classes such as Starfish and Holothurian. The +Residual Attention model achieved the highest mAP of 57.09, emphasizing the value of attention mechanisms. However, +Channel Attention+Multi-Scale Attention achieved the highest mAP of 55.57 and excelled in Echinus and Holothurian, demonstrating the potential of multi-attention models.

Table \ref{tab:results_augmented_no_domain} evaluates models on augmented datasets without domain-specific information. The +Residual Attention model achieved the highest mAP of 34.78, particularly excelling in the Holothurian class. However, the +Residual+Channel Attention+Multi-Scale Attention model outperformed others with an mAP of 54.62, highlighting the value of combining multiple attention mechanisms in diverse scenarios.

In Table \ref{tab:results_original_with_domain}, we observe the impact of domain-specific information on model performance with the original dataset. Domain knowledge significantly enhances detection, especially in classes like Starfish and Holothurian. The +Domain +Residual+Multi-Scale Attention model achieved the highest mAP of 58.57, emphasizing the advantage of combining domain-specific information with multi-attention mechanisms.

Table \ref{tab:results_augmented_with_domain} assesses model performance on augmented datasets with domain-specific information. Notably, +Domain +Residual+Multi-Scale Attention emerged as the top performer with an mAP of 36.16, significantly improving detection in Holothurian and Scallop classes.

In summary, our analysis highlights the need for tailored approaches to specific classes, the effectiveness of multi-attention models, and the substantial benefits of incorporating domain knowledge, whether working with original or augmented datasets. The choice of the best-performing model should consider the unique detection requirements and contextual nuances of each class.

Figure \ref{fig:1} offers a visual summary of the detection results obtained using the original dataset. In the top row, subfigures correspond to models trained without the use of domain-specific information. These models can be directly related to the information presented in Table 1, which allows us to discern the best and worst performers at a glance.

The observations from Figure \ref{fig:1} align with the tabular data in Table 1. For instance, the subfigure denoted as Model 4, represents the model that incorporates residual attention and consistently outperforms other models in terms of mean Average Precision (mAP). Conversely, subfigure corresponds to the baseline model, YOLOv3, which serves as a reference point for performance comparison.

In the bottom row of Figure \ref{fig:1}, subfigures portray models that integrate domain-specific knowledge. Again, these models mirror the information in Table 1, and we can directly associate their performance with the tabular data. Subfigure labeled as Model 8, clearly stands out as the top performer. This reinforces the significance of domain-specific information for enhancing object detection accuracy.

Figure \ref{fig:2} extends the evaluation to models trained on the augmented dataset (Type 8), which introduces additional complexities compared to the original dataset. Similar to Figure \ref{fig:1}, this figure provides an easy-to-understand visual representation of detection results.

In the top row, subfigures represent models trained without domain-specific information using the augmented dataset. These subfigures are in correspondence with the data in Table 2, allowing us to swiftly identify the best and worst-performing models. Subfigure of Model 4, which employs residual attention, consistently achieves the highest mAP, aligning with the insights from Table 2.

In the bottom row of Figure \ref{fig:2}, subfigures (i) to (p) portray models that incorporate domain-specific information with the augmented dataset. Again, the sub-captions Model 1 to Model 8 correspond to the tabular data. Subfigure labeled as Model 4, continues to demonstrate the best performance, reinforcing the importance of domain-specific knowledge even in the presence of augmented and more complex datasets.

In summary, Figures \ref{fig:1} and \ref{fig:2}, with their sub-captions referencing Model 1 to Model 8, offer a clear and concise representation of model performance under various conditions. These visuals complement the tabular data (Table 1 and Table 2) and facilitate a quick and intuitive understanding of the best detection outcomes across different models and datasets, which is essential for decision-making in our domain-specific application.

The MARS model exhibits promising results in underwater object detection; however, several limitations require careful consideration:
\begin{itemize}

\item  \textbf { Domain-Specific Training Data:} MARS's performance heavily relies on domain-specific training data. It may face challenges when applied to entirely new and uncharted underwater domains. Future research should emphasize collecting data from diverse underwater environments to enhance the model's adaptability.

\item  \textbf { Resource Intensiveness:} The integration of complex modules like Residual Attention and Domain-Adaptive Multi-Scale Attention (DAMSA) demands significant computational resources. This can hinder deployment on resource-constrained underwater robotic platforms. To make MARS more accessible, research should focus on model compression and hardware acceleration techniques.

\item  \textbf { Extreme Environmental Conditions:} MARS may encounter limitations when operating in extreme underwater conditions, such as extreme depths, turbid waters, or harsh weather. Investigating adaptive algorithms and robust sensor technologies is crucial to improve its performance in these conditions.

\item \textbf{Generalization to New Object Classes:} While MARS excels in detecting specific underwater object classes, its effectiveness in recognizing new classes or other underwater items needs further exploration. Ongoing research should aim to improve the model's ability to generalize to novel classes, expanding its detection capabilities. To achieve this, we plan to conduct extensive testing of MARS using real-world underwater datasets collected from diverse ocean environments, ensuring its adaptability to a wide range of underwater objects and conditions.

\item  \textbf { Interpretability and Explainability:} MARS, like many deep learning models, can be challenging to interpret. This lack of interpretability can be a barrier to trust and collaboration with marine researchers. Future work should delve into explainable AI (XAI) techniques tailored to underwater object detection to enhance the model's transparency.
\end{itemize}

To address these limitations, future work should involve interdisciplinary collaborations with marine scientists and roboticists, explore model variants, and invest in research areas like data collection, model compression, and underwater sensor development. By addressing these challenges, we can advance underwater object detection and enhance the practical applicability of the MARS model in real-world scenarios.

\section{CONCLUSIONS}

In this study, we introduced MARS (Multi-Scale Adaptive Robotics Vision), a novel approach to underwater object detection designed to excel in diverse underwater scenarios. MARS leverages innovations built upon the YOLOv3 framework, integrating advanced techniques like residual layers, domain adaptation, channel attention, and multi-scale attention modules to significantly elevate detection performance.
On the original dataset, MARS exhibits exceptional performance, achieving a remarkable mean Average Precision (mAP) of 58.57\%. This result stands as a testament to MARS's proficiency in detecting essential underwater objects, including echinus, starfish, holothurian, scallop, and waterweeds. This capability holds immense promise for applications in marine robotics, marine biology research, and environmental monitoring.
Furthermore, MARS excels at addressing the challenge of domain shifts. On the augmented dataset, with the inclusion of all enhancements (+Domain +Residual+Channel Attention+Multi-Scale Attention), MARS achieves an outstanding mAP of 36.16\%. This result not only reaffirms its robustness but also underscores its adaptability in recognizing objects amidst previously unencountered underwater conditions.

\section*{Acknowledgement}
\noindent This work is supported by the Khalifa University of Science and Technology under Award No.  RC1-2018-KUCARS.


\bibliographystyle{IEEEtran}
\balance
\bibliography{IEEEabrv,references}

\begin{thebibliography}{10}
\providecommand{\url}[1]{#1}
\csname url@rmstyle\endcsname
\providecommand{\newblock}{\relax}
\providecommand{\bibinfo}[2]{#2}
\providecommand\BIBentrySTDinterwordspacing{\spaceskip=0pt\relax}
\providecommand\BIBentryALTinterwordstretchfactor{4}
\providecommand\BIBentryALTinterwordspacing{\spaceskip=\fontdimen2\font plus
\BIBentryALTinterwordstretchfactor\fontdimen3\font minus
  \fontdimen4\font\relax}
\providecommand\BIBforeignlanguage[2]{{%
\expandafter\ifx\csname l@#1\endcsname\relax
\typeout{** WARNING: IEEEtran.bst: No hyphenation pattern has been}%
\typeout{** loaded for the language `#1'. Using the pattern for}%
\typeout{** the default language instead.}%
\else
\language=\csname l@#1\endcsname
\fi
#2}}

\bibitem{recht2019imagenet-1}
B.~Recht, R.~Roelofs, L.~Schmidt, and V.~Shankar, ``Do imagenet classifiers
  generalize to imagenet?'' in \emph{International conference on machine
  learning}.\hskip 1em plus 0.5em minus 0.4em\relax PMLR, 2019, pp. 5389--5400.

\bibitem{hendrycks2019benchmarking-2}
D.~Hendrycks and T.~Dietterich, ``Benchmarking neural network robustness to
  common corruptions and perturbations,'' \emph{arXiv preprint
  arXiv:1903.12261}, 2019.

\bibitem{chen2018domain-3}
Y.~Chen, W.~Li, C.~Sakaridis, D.~Dai, and L.~Van~Gool, ``Domain adaptive faster
  r-cnn for object detection in the wild,'' in \emph{Proceedings of the IEEE
  conference on computer vision and pattern recognition}, 2018, pp. 3339--3348.

\bibitem{xu2020exploring-4}
C.-D. Xu, X.-R. Zhao, X.~Jin, and X.-S. Wei, ``Exploring categorical
  regularization for domain adaptive object detection,'' in \emph{Proceedings
  of the IEEE/CVF Conference on Computer Vision and Pattern Recognition}, 2020,
  pp. 11\,724--11\,733.

\bibitem{hsu2020progressive-5}
H.-K. Hsu, C.-H. Yao, Y.-H. Tsai, W.-C. Hung, H.-Y. Tseng, M.~Singh, and M.-H.
  Yang, ``Progressive domain adaptation for object detection,'' in
  \emph{Proceedings of the IEEE/CVF winter conference on applications of
  computer vision}, 2020, pp. 749--757.

\bibitem{li2017deeper-6}
D.~Li, Y.~Yang, Y.-Z. Song, and T.~M. Hospedales, ``Deeper, broader and artier
  domain generalization,'' in \emph{Proceedings of the IEEE international
  conference on computer vision}, 2017, pp. 5542--5550.

\bibitem{motiian2017unified}
S.~Motiian, M.~Piccirilli, D.~Adjeroh, and G.~Doretto, ``Unified deep
  supervised domain adaptation and generalization,'' in \emph{Proc. IEEE Int.
  Conf. Comput. Vision}, 2017, pp. 5715--5725.

\bibitem{li2018domain}
H.~Li, S.~Jialin~Pan, S.~Wang, and A.~Kot, ``Domain generalization with
  adversarial feature learning,'' in \emph{Proc. IEEE Conf. Comput. Vision
  Pattern Recognit.}, 2018, pp. 5400--5409.

\bibitem{shankar2018generalizing}
S.~Shankar, V.~Piratla, S.~Chakrabarti, S.~Chaudhuri, P.~Jyothi, and
  S.~Sarawagi, ``Generalizing across domains via cross-gradient training,'' in
  \emph{Proc. Int. Conf. Learn. Repre}, 2018.

\bibitem{dou2019domain}
Q.~Dou, D.~Castro, K.~Kamnitsas, and B.~Glocker, ``Domain generalization via
  model-agnostic learning of semantic features,'' in \emph{Proc. Conf. Neur.
  Info. Proc. Systems}, 2019, pp. 6447--6458.

\bibitem{d2018domain}
A.~D’Innocente and B.~Caputo, ``Domain generalization with domain-specific
  aggregation modules,'' in \emph{Proc. Germ. Conf. Pattern Recognit.}\hskip
  1em plus 0.5em minus 0.4em\relax Springer, 2018, pp. 187--198.

\bibitem{balaji2018metareg}
Y.~Balaji, S.~Sankaranarayanan, and R.~Chellappa, ``Metareg: Towards domain
  generalization using meta-regularization,'' in \emph{Proc. Conf. Neur. Info.
  Proc. Systems}, 2018, pp. 998--1008.

\bibitem{li2019episodic}
D.~Li, J.~Zhang, Y.~Yang, C.~Liu, Y.~Song, and T.~Hospedales, ``Episodic
  training for domain generalization,'' \emph{arXiv:1902.00113}, 2019.

\bibitem{huang2019faster}
H.~Huang, H.~Zhou, X.~Yang, L.~Zhang, L.~Qi, and A.~Zang, ``Faster r-cnn for
  marine organisms detection and recognition using data augmentation,''
  \emph{Neurocomputing}, vol. 337, pp. 372--384, 2019.

\bibitem{lin2020roimix}
W.~Lin, J.~Zhong, S.~Liu, T.~Li, and G.~Li, ``Roimix: Proposal-fusion among
  multiple images for underwater object detection,'' in \emph{IEEE
  International Conference on Acoustics, Speech, and Signal Processing}, 2020.

\bibitem{liu2020towards}
H.~Liu, P.~Song, and R.~Ding, ``Towards domain generalization in underwater
  object detection,'' in \emph{Proc. Int. Conf. Image Proc.}, 2020, pp.
  1971--1975.

\bibitem{CHEN202320}
\BIBentryALTinterwordspacing
Y.~Chen, P.~Song, H.~Liu, L.~Dai, X.~Zhang, R.~Ding, and S.~Li, ``Achieving
  domain generalization for underwater object detection by domain mixup and
  contrastive learning,'' \emph{Neurocomputing}, vol. 528, pp. 20--34, 2023.
  [Online]. Available:
  \url{https://www.sciencedirect.com/science/article/pii/S0925231223000644}
\BIBentrySTDinterwordspacing

\bibitem{fan2020dual}
B.~Fan, W.~Chen, Y.~Cong, and J.~Tian, ``Dual refinement underwater object
  detection network,'' in \emph{Computer Vision -- ECCV 2020}, A.~Vedaldi,
  H.~Bischof, T.~Brox, and J.-M. Frahm, Eds.\hskip 1em plus 0.5em minus
  0.4em\relax Cham: Springer International Publishing, 2020, pp. 275--291.

\bibitem{chen2020c}
X.~Chen, Y.~Lu, Z.~Wu, J.~Yu, and L.~Wen, ``Reveal of domain effect: How visual
  restoration contributes to object detection in aquatic scenes,'' \emph{arXiv
  preprint arXiv:2003.01913}, 2020.

\bibitem{liang2022excavating}
X.~Liang and P.~Song, ``Excavating roi attention for underwater object
  detection.''\hskip 1em plus 0.5em minus 0.4em\relax IEEE, 2022.

\bibitem{zhao2021composited}
Z.~Zhao, Y.~Liu, X.~Sun, J.~Liu, X.~Yang, and C.~Zhou, ``Composited fishnet:
  Fish detection and species recognition from low-quality underwater videos,''
  \emph{IEEE Transaction on Image Processing}, vol.~30, pp. 4719--4734, 2021.

\bibitem{ganin2015unsupervised}
Y.~Ganin and V.~Lempitsky, ``Unsupervised domain adaptation by
  backpropagation,'' in \emph{Proc. Int. Conf. Mach. Learn.}, 2015, pp.
  1180--1189.

\bibitem{xu2019adversarial}
M.~Xu, J.~Zhang, B.~Ni, T.~Li, C.~Wang, Q.~Tian, and W.~Zhang, ``Adversarial
  domain adaptation with domain mixup,'' in \emph{Proc. AAAI Conf. Arti.
  Intell.}, 2019, pp. 6502--6509.

\bibitem{gong2019dlow}
R.~Gong, W.~Li, Y.~Chen, and L.~Van~Gool, ``Dlow: Domain flow for adaptation
  and generalization,'' in \emph{Proc. IEEE Conf. Comput. Vision Pattern
  Recognit.}, 2019, pp. 2477--2486.

\bibitem{chen2018domain}
Y.~Chen, W.~Li, C.~Sakaridis, D.~Dai, and L.~Van~Gool, ``Domain adaptive faster
  r-cnn for object detection in the wild,'' in \emph{Proc. IEEE Conf. Comput.
  Vision Pattern Recognit.}, 2018, pp. 3339--3348.

\bibitem{10161439}
H.~Liu, M.~Roznere, and A.~Q. Li, ``Deep underwater monocular depth estimation
  with single-beam echosounder,'' in \emph{2023 IEEE International Conference
  on Robotics and Automation (ICRA)}, 2023, pp. 1090--1097.

\bibitem{10160331}
S.~Chen, H.~Xu, X.~Xiong, and B.~Lu, ``An underwater jet-propulsion soft robot
  with high flexibility driven by water hydraulics,'' in \emph{2023 IEEE
  International Conference on Robotics and Automation (ICRA)}, 2023, pp.
  2613--2619.

\bibitem{10160853}
K.~Yao, N.~Bauschmann, T.~L. Alff, W.~Cheah, D.~A. Duecker, K.~Groves,
  O.~Marjanovic, and S.~Watson, ``Image-based visual servoing switchable
  leader-follower control of heterogeneous multi-agent underwater robot
  system,'' in \emph{2023 IEEE International Conference on Robotics and
  Automation (ICRA)}, 2023, pp. 5200--5206.

\bibitem{10161282}
Y.~Girdhar, N.~McGuire, L.~Cai, S.~Jamieson, S.~McCammon, B.~Claus, J.~E.~S.
  Soucie, J.~E. Todd, and T.~A. Mooney, ``Curee: A curious underwater robot for
  ecosystem exploration,'' in \emph{2023 IEEE International Conference on
  Robotics and Automation (ICRA)}, 2023, pp. 11\,411--11\,417.

\bibitem{redmon2018yolov3}
J.~Redmon and A.~Farhadi, ``{YOLOv3}: An incremental improvement,'' \emph{arXiv
  preprint arXiv:1804.02767}, 2018.

\end{thebibliography}

\end{document}